\documentclass{article}

\usepackage[paperwidth=8.5in,paperheight=11.0in,
  left=1.0in,right=1.0in,top=1.0in,bottom=1.0in,
  heightrounded]{geometry}

\usepackage{sectsty}
\usepackage{amsmath}
\sectionfont{\normalsize}
\subsectionfont{\normalsize}

\usepackage[numbers]{natbib}
\title{\textbf{Why AI is Harder Than We Think}}
\author{\textbf{Melanie Mitchell} \\ \normalsize Santa Fe Institute \\ Santa Fe, NM, USA \\ mm@santafe.edu \\
}
\date{}
\begin{document}
\maketitle

\begin{center} \textbf{Abstract} \end{center}
Since its beginning in the 1950s, the field of artificial intelligence has cycled several times between periods of optimistic predictions and massive investment (``AI spring'') and periods of disappointment, loss of confidence, and reduced funding (``AI winter''). Even with today’s seemingly fast pace of AI breakthroughs, the development of long-promised technologies such as self-driving cars, housekeeping robots, and conversational companions has turned out to be much harder than many people expected. One reason for these repeating cycles is our limited understanding of the nature and complexity of intelligence itself. In this paper I describe four fallacies in common assumptions made by AI researchers, which can lead to overconfident predictions about the field. I conclude by discussing the open questions spurred by these fallacies, including the age-old challenge of imbuing machines with humanlike common sense.

\section*{Introduction}
The year 2020 was supposed to herald the arrival of self-driving cars.  Five years earlier, a headline in \textit{The Guardian} predicted that ``From 2020 you will become a permanent backseat driver'' \cite{Guardian2015}. In 2016 \textit{Business Insider} assured us that ``10 million self-driving cars will be on the road by 2020'' \cite{BusinessInsider2016}.  Tesla Motors CEO Elon Musk promised in 2019 that ``A year from now, we'll have over a million cars with full self-driving, software...everything'' \cite{Verge2019}. And 2020 was the target announced by several automobile companies to bring self-driving cars to market \cite{Verge2017,PhysOrg2015,Toyota2020}.

Despite attempts to redefine ``full self-driving'' into existence \cite{CarAndDriver2021}, none of these predictions has come true.  It's worth quoting AI expert Drew McDermott on what can happen when over-optimism about AI systems---in particular, self-driving cars---turns out to be wrong:
\begin{quote}
   Perhaps expectations are too high, and... this will eventually result in disaster. [S]uppose that five years from now [funding] collapses miserably as autonomous vehicles fail to roll.  Every startup company fails. And there's a big backlash so that you can't get money for anything connected with AI. Everybody hurriedly changes the names of their research projects to something else. This condition [is] called the ``AI Winter'' \cite{McDermott1985}.
\end{quote}

What's most notable is that McDermott's warning is from 1984, when, like today, the field of AI was awash with confident optimism about the near future of machine intelligence. McDermott was writing about a cyclical pattern in the field.  New, apparent breakthroughs would lead AI practitioners to predict rapid progress, successful commercialization, and the near-term prospects of ``true AI.'' Governments and companies would get caught up in the enthusiasm, and would shower the field with research and development funding.  AI Spring would be in bloom.  When progress stalled, the enthusiasm, funding, and jobs would dry up. AI Winter would arrive.  Indeed, about five years after McDermott's warning, a new AI winter set in.

In this chapter I explore the reasons for the repeating cycle of overconfidence followed by disappointment in expectations about AI.  I argue that over-optimism among the public, the media, and even experts can arise from several fallacies in how we talk about AI and in our intuitions about the nature of intelligence.  Understanding these fallacies and their subtle influences can point to directions for creating more robust, trustworthy, and perhaps actually \textit{intelligent} AI systems.

\section*{Springs and Winters}
Overconfident predictions about AI are as old as the field itself.  In 1958, for example, the New York Times reported on a demonstration by the US Navy of Frank Rosenblatt's ``perceptron'' (a rudimentary precursor to today's deep neural networks): ``The Navy revealed the embryo of an electronic computer today that it expects will be able to walk, talk, see, write, reproduce itself, and be conscious of its existence'' \cite{NYT1958}.  This optimistic take was quickly followed by similar proclamations from AI pioneers, this time about the promise of logic-based ``symbolic'' AI.  In 1960 Herbert Simon declared that, ``Machines will be capable, within twenty years, of doing any work that a man can do'' \cite{Simon1960}.  The following year, Claude Shannon echoed this prediction: ``I confidently expect that within a matter of 10 or 15 years, something will emerge from the laboratory which is not too far from the robot of science fiction fame'' \cite{Shannon1961}. And a few years later Marvin Minsky forecast that, ``Within a generation...the problems of creating ‘artificial intelligence' will be substantially solved'' \cite{Minsky1967}.

The optimistic AI Spring of the 1960s and early 1970s, reflected in these predictions, soon gave way to the first AI winter.  Minsky and Papert's 1969 book Perceptrons \cite{Minsky1969} showed that the kinds of problems solvable by Rosenblatt's perceptrons were very limited. In 1973 the Lighthill report \cite{Lighthill1973} in the UK and the Department of Defense's ``American Study Group'' report in the US, commissioned by their respective governments to assess prospects for AI in the near future, were both extremely negative about those prospects. This led to sharp funding decreases and a downturn in enthusiasm for AI in both countries.

AI once again experienced an upturn in enthusiasm starting in the early 1980s with several new initiatives: the rise of ``expert systems'' in industry \cite{Durkin1996},  Japan's huge investment in its ``Fifth Generation'' project \cite{Gaines1984}, which aimed for ambitious AI abilities as the core of a new generation of computing systems, the US's responding ``Strategic Computing Initiative'' \cite{Stefik1985}, which provided large funding for progress into general AI, as well as a new set of efforts on neural networks \cite{McClelland1986a,McClelland1986b}, which generated new hopes for the field.

By the latter part of the 1980s, these optimistic hopes had all been dashed; again, none of these technologies had achieved the lofty promises that had been made.  Expert systems, which rely on humans to create rules that capture expert knowledge of a particular domain, turned out to be brittle---that is, often unable to generalize or adapt when faced with new situations.  The problem was that the human experts writing the rules actually rely on subconscious knowledge---what we might call ``common sense''---that was not part of the system's programming.  The AI approaches pursued under the Fifth Generation project and Strategic Computing Initiative ran into similar problems of brittleness and lack of generality.  The neural-network approaches of the 1980s and 1990s likewise worked well on relatively simple examples but lacked the ability to scale up to complex problems. Indeed, the late 1980's marked the beginning of a new AI winter, and the field's reputation suffered.  When I received my PhD in 1990, I was advised not to use the term ``artificial intelligence'' on my job applications.

At the 50th anniversary commemoration of the 1956 Dartmouth Summer Workshop that launched the field, AI pioneer John McCarthy, who had originally coined the term ``Artificial Intelligence,'' explained the issue succinctly: ``AI was harder than we thought'' \cite{Moewes2013}.

The 1990s and 2000s saw the meteoric rise of \textit{machine learning}: the development of algorithms that create predictive models from data.  These approaches were typically inspired by statistics rather than by neuroscience or psychology, and were aimed at performing specific tasks rather than capturing general intelligence.  Machine-learning practitioners were often quick to differentiate their discipline from the then-discredited field of AI.  

However, around 2010, \textit{deep learning}---in which brain-inspired multilayered neural networks are trained from data---emerged from its backwater position and rose to superstar status in machine learning. Deep neural networks had been around since the 1970s, but only recently, due to huge datasets scraped from the Web, fast parallel computing chips, and innovations in training methods, could these methods scale up to a large number of previously unsolved AI challenges.  Deep neural networks are what power all of the major AI advances we've seen in the past decade, including speech recognition, machine translation, chat bots, image recognition, game playing, and protein folding, among others.

Suddenly the term ``AI'' started to appear everywhere, and there was all at once a new round of optimism about the prospects of what has been variously called ``general,'' ``true,'' or ``human-level'' AI.   

In surveys of AI researchers carried out in 2016 and 2018, the median prediction of those surveyed gave a 50 percent chance that human-level AI would be created by 2040–2060, though there was much variance of opinion, both for sooner and later estimates \cite{Muller2016,Grace2018}. Even some of the most well-known AI experts and entrepreneurs are in accord.  Stuart Russell, co-author of a widely used textbook on AI, predicts that ``superintelligent AI'' will ``probably happen in the lifetime of my children'' \cite{Russell2019} and Sam Altman, CEO of the AI company OpenAI, predicts that within decades, computer programs ``will do almost everything, including making new scientific discoveries that will expand our concept of `everything.'\thinspace'' \cite{Altman2021} Shane Legg, co-founder of Google DeepMind, predicted in 2008 that, ``Human level AI will be passed in the mid-2020s'' \cite{Despres2008}, and Facebook's CEO, Mark Zuckerberg, declared in 2015 that ``One of [Facebook's] goals for the next five to 10 years is to basically get better than human level at all of the primary human senses: vision, hearing, language, general cognition'' \cite{McCracken2015}.

However, in spite of all the optimism, it didn't take long for cracks to appear in deep learning's fa\c{c}ade of intelligence. It turns out that, like all AI systems of the past, deep-learning systems can exhibit brittleness---unpredictable errors when facing situations that differ from the training data.  This is because such systems are susceptible to \textit{shortcut learning} \cite{Geirhos2020,Lapuschkin2019}: learning statistical associations in the training data that allow the machine to produce correct answers but sometimes for the wrong reasons.  In other words, these machines  don't learn the concepts we are trying to teach them, but rather they learn shortcuts to correct answers on the training set---and such shortcuts will not lead to good generalizations.  Indeed, deep learning systems often cannot learn the abstract concepts that would enable them to transfer what they have learned to new situations or tasks \cite{Mitchell2021}. Moreover, such systems are vulnerable to attack from ``adversarial perturbations'' \cite{Moosavi2017}---specially engineered changes to the input that are either imperceptible or irrelevant to humans, but that induce the system to make errors.

Despite extensive research on the limitations of deep neural networks, the sources of their brittleness and vulnerability are still not completely understood. These  networks, with their large number of parameters, are complicated systems whose decision-making mechanisms can be quite opaque. However, it seems clear from their non-humanlike errors and vulnerability to adversarial perturbations that these systems are not actually \textit{understanding} the data they process, at least not in the human sense of ``understand.''  It's still a matter of debate in the AI community whether such understanding can be achieved by adding network layers and more training data, or whether something more fundamental is missing.

At the time of this writing (mid-2021), several new deep-learning approaches are once again generating considerable optimism in the AI community.  Some of the hottest new areas are transformer architectures using self-supervised (or ``predictive'') learning \cite{Devlin2018}, meta-learning \cite{Finn2017}, and deep reinforcement learning \cite{Arulkumaran2017}; each of these has been cited as progress towards more general, human-like AI. While these and other new innovations have shown preliminary promise, the AI cycle of springs and winters is likely to continue.  The field continually advances in relatively narrow areas, but the path toward human-level AI is less clear.  

In the next sections I will argue that predictions about the likely timeline of human-level AI reflect our own biases and lack of understanding of the nature of intelligence.  In particular, I describe four fallacies in our thinking about AI that seem most central to me. While these fallacies are well-known in the AI community, many assumptions made by experts still fall victim to these fallacies, and give us a false sense of confidence about the near-term prospects of ``truly'' intelligent machines.

\section*{Fallacy 1: Narrow intelligence is on a continuum with general intelligence}
Advances on a specific AI task are often described as ``a first step'' towards more general AI.  The chess-playing computer Deep Blue was ``was hailed as the first step of an AI revolution'' \cite{NewScientist2016}. IBM described its Watson system as ``a first step into cognitive systems, a new era of computing'' \cite{IBM2013}.  OpenAI's GPT-3 language generator was called a ``step toward general intelligence'' \cite{SSC2019}. 

Indeed, if people see a machine do something amazing, albeit in a narrow area, they often assume the field is that much further along toward general AI.  The philosopher Hubert Dreyfus (using a term coined by Yehoshua Bar-Hillel) called this a ``first-step fallacy.'' As Dreyfus characterized it, ``The first-step fallacy is the claim that, ever since our first work on computer intelligence we have been inching along a continuum at the end of which is AI so that any improvement in our programs no matter how trivial counts as progress.''   Dreyfus quotes an analogy made by his brother, the engineer Stuart Dreyfus: ``It was like claiming that the first monkey that climbed a tree was making progress towards landing on the moon'' \cite{Dreyfus2012}.

Like many AI experts before and after him, Dreyfus noted that the ``unexpected obstacle'' in the assumed continuum of AI progress has always been the problem of \textit{common sense}.  I will say more about this barrier of common sense in the last section.  

\section*{Fallacy 2: Easy things are easy and hard things are hard}
While John McCarthy lamented that ``AI was harder than we thought,'' Marvin Minsky explained that this is because ``easy things are hard'' \cite{Minsky1987}.  That is, the things that we humans do without much thought---looking out in the world and making sense of what we see, carrying on a conversation, walking down a crowded sidewalk without bumping into anyone---turn out to be the hardest challenges for machines. Conversely, it's often easier to get machines to do things that are very hard for humans; for example, solving complex mathematical problems, mastering games like chess and Go, and translating sentences between hundreds of languages have all turned out to be relatively easier for machines.  This is a form of what's been called ``Moravec's paradox,'' named after roboticist Hans Moravec, who wrote, ``It is comparatively easy to make computers exhibit adult level performance on intelligence tests or playing checkers, and difficult or impossible to give them the skills of a one-year-old when it comes to perception and mobility'' \cite{Moravec1988a}.

This fallacy has influenced thinking about AI since the dawn of the field. AI pioneer Herbert Simon proclaimed that ``Everything of interest in cognition happens above the 100-millisecond level---the time it takes you to recognize your mother'' \cite{Hofstadter1985}.  Simon is saying that, to understand cognition, we don't have to worry about unconscious perceptual processes.  This assumption is reflected in most of the symbolic AI tradition, which focuses on the process of reasoning about input that has already been perceived. 

In the last decades, symbolic AI approaches have lost favor in the research community, which has largely been dominated by deep learning, which does address perception.  However, the assumptions underlying this fallacy still appear in recent claims about AI.  For example, in a 2016 article, deep-learning pioneer Andrew Ng was quoted echoing Simon's assumptions, vastly underestimating the complexity of unconscious perception and thought: ``If a typical person can do a mental task with less than one second of thought, we can probably automate it using AI either now or in the near future'' \cite{Ng2016}.

More subtly, researchers at Google DeepMind, in talking about AlphaGo's triumph, described the game of Go as one of ``the most challenging of domains'' \cite{Silver2017}.  Challenging for whom?  For humans, perhaps, but as psychologist Gary Marcus pointed out, there are domains, including games, that, while easy for humans, are much more challenging than Go for AI systems.  One example is charades, which ``requires acting skills, linguistic skills, and theory of mind'' \cite{Marcus2018}, abilities that are far beyond anything AI can accomplish today. 

AI is harder than we think, because we are largely unconscious of the complexity of our own thought processes. Hans Moravec explains his paradox this way: ``Encoded in the large, highly evolved sensory and motor portions of the human brain is a billion years of experience about the nature of the world and how to survive in it. The deliberate process we call reasoning is, I believe, the thinnest veneer of human thought, effective only because it is supported by this much older and much more powerful, though usually unconscious, sensorimotor knowledge. We are all prodigious Olympians in perceptual and motor areas, so good that we make the difficult look easy'' \cite{Moravec1988b}. Or more succinctly, Marvin Minsky notes, ``In general, we're least aware of what our minds do best'' \cite{Minsky1980}. 

\section*{Fallacy 3: The lure of wishful mnemonics}
The term ``wishful mnemonic'' was coined in a 1976 critique of AI by computer scientist Drew McDermott:

\begin{quote}
     A major source of simple-mindedness in AI programs is the use of mnemonics like ``UNDERSTAND'' or ``GOAL'' to refer to programs and data structures. ...If a researcher...calls the main loop of his program ``UNDERSTAND,'' he is (until proven innocent) merely begging the question. He may mislead a lot of people, most prominently himself. ...What he should do instead is refer to this main loop as ``G0034,'' and see if he can convince himself or anyone else that G0034 implements some part of understanding. ...Many instructive examples of wishful mnemonics by AI researchers come to mind once you see the point \cite{McDermott1976}.
\end{quote}

Now, many decades later, work on AI is replete with such wishful mnemonics---terms associated with human intelligence that are used to describe the behavior and evaluation of AI programs.  \textit{Neural} networks are loosely inspired by the brain, but with vast differences.  Machine \textit{learning} or deep \textit{learning} methods do not really resemble learning in humans (or in non-human animals).  Indeed, if a machine has learned something in the human sense of \textit{learn}, we would expect that it would be able use what it has learned in different contexts.  However, it turns out that this is often not the case.  In machine learning there is an entire subfield called \textit{transfer learning} that focuses on the still-open problem of how to enable machines to transfer what they have learned to new situations, an ability that is fundamental to human learning.

Indeed, the way we talk about machine abilities influences our conceptions of how general those abilities really are.  Unintentionally providing real-world illustrations of McDermott's warning, one of IBM's top executives proclaimed that ``Watson can \textit{read} all of the health-care texts in the world in seconds'' \cite{Gustin2011} and IBM's website claims that its Watson program ``\textit{understands} context and nuance in seven languages'' \cite{IBMCognitive}. DeepMind co-founder Demis Hassabis tells us that ``AlphaGo's \textit{goal} is to beat the best human players not just mimic them'' \cite{KoreaHerald2016}. And AlphaGo's lead research David Silver described one of the program's matches thus: ``We can always ask AlphaGo how well it \textit{thinks} it's doing during the game. ...It was only towards the end of the game that \textit{AlphaGo thought it would win}''  \cite{Shead2017}. (Emphasis is mine in the quotations above.) 

One could argue that such anthropomorphic terms are simply shorthand: IBM scientists know that Watson doesn't read or understand in the way humans do; DeepMind scientists know that AlphaGo has no goals or thoughts in the way humans do, and no human-like conceptions of a ``game'' or of ``winning.'' However, such shorthand can be misleading to the public trying to understand these results (and to the media reporting on them), and can also unconsciously shape the way even AI experts think about their systems and how closely these systems resemble human intelligence.

McDermott's ``wishful mnemonics'' referred to terms we use to describe AI programs, but the research community also uses wishful mnemonics in naming AI evaluation benchmarks after the skills we hope they test.  For example, here are some of the most widely cited current benchmarks in the subarea of AI called ``natural-language processing'' (NLP):  the ``Stanford Question Answering Dataset'' \cite{SQUAD}, the ``RACE Reading Comprehension Dataset'' \cite{RACEDataset}, and the ``General Language Understanding Evaluation'' \cite{GLUEBenchmark}.   In all of these benchmarks, the performance of the best machines has already exceeded that measured for humans (typically Amazon Mechanical Turk workers).  This has led to headlines such as ``New AI model exceeds human performance at question Answering'' \cite{Costenaro2018}; ``Computers are getting better than humans at reading'' \cite{Pham2018};  and ``Microsoft's AI model has outperformed humans in natural-language understanding'' \cite{Jawad2021}. Given the names of these benchmark evaluations, it's not surprising that people would draw such conclusions.  The problem is, these benchmarks don't actually measure general abilities for question-answering, reading comprehension, or natural-language understanding. The benchmarks test only very limited versions of these abilities; moreover, many of these benchmarks allow machines to learn shortcuts, as I described above---statistical correlations that machines can exploit to achieve high performance on the test without learning the actual skill being tested \cite{McCoy2019,Linzen2020}. While machines can outperform humans on these particular benchmarks, AI systems are still far from matching the more general human abilities we associate with the benchmarks' names.  

\section*{Fallacy 4: Intelligence is all in the brain}

The idea that intelligence is something that can be separated from the body, whether as a non-physical substance or as wholly encapsulated in the brain, has a long history in philosophy and cognitive science. 

The so-called ``information-processing model of mind'' arose in psychology in the mid-twentieth century.  This model views the mind as a kind of computer, which inputs, stores, processes, and outputs information. The body does not play much of a role except in the input (perception) and output (behavior) stages.  Under this view, cognition takes place wholly in the brain, and is, in theory, separable from the rest of the body.  An extreme corollary of this view is that, in the future, we will be able to ``upload'' our brains---and thus our cognition and consciousness---to computers \cite{Woollaston2013}. 

The assumption that intelligence can in principle be ``disembodied'' is implicit in almost all work on AI throughout its history.  One of the most influential ideas in early AI research was Newell and Simon's ``Physical Symbol System Hypothesis'' (PSSH), which stated: ``A physical symbol system has the necessary and sufficient means for general intelligent action'' \cite{Newell1976}. The term ``physical symbol system'' refers to something much like a digital computer.  The PSSH posits that general intelligence can be achieved in digital computers without incorporating any non-symbolic processes of brain or body.  (For an insightful discussion of symbolic versus subsymbolic processes, see Hofstadter's ``Waking up from the Boolean Dream'' \cite{Hofstadter1985b}.)

Newell and Simon's PSSH was a founding principle of the symbolic approach to AI, which dominated the field until the rise of statistical and neurally inspired machine learning in the 1990s and 2000s.  However, these latter non-symbolic approaches also did not view the body as relevant to intelligence.  Instead, neurally inspired approaches from 1980s connectionism to today's deep neural networks generally assume that intelligence arises solely from brain structures and dynamics.  Today's deep neural networks are akin to the proverbial brain-in-a-vat: passively taking in data from the world and outputting instructions for behavior without actively interacting in the world with any kind of body.  Of course, robots and autonomous vehicles are different in that they have a physical presence in the world, but to date the kinds of physical interactions they have, and the feedback to their ``intelligence'' is quite limited. 

The assumption that intelligence is all in the brain has led to speculation that, to achieve human-level AI, we simply need to scale up machines to match the brain's ``computing capacity'' and then develop the appropriate ``software'' for this brain-matching ``hardware.  For example, one philosopher wrote a report on the literature that concluded, ``I think it more likely than not that $10^{15}$ FLOP/s is enough to perform tasks as well as the human brain (given the right software, which may be very hard to create)'' \cite{Carlsmith2020}.  No body needed!

Top AI researchers have echoed the idea that scaling up hardware to match the brain will enable human-level artificial intelligence.  For example, deep-learning pioneer Geoffrey Hinton predicted, ``To understand [documents] at a human level, we're probably going to need human-level resources and we have trillions of connections [in our brains]. ...But the biggest networks we have built so far only have billions of connections. So we're a few orders of magnitude off, but I'm sure the hardware people will fix that'' \cite{Patterson2017}. Others have predicted that the ``hardware fix''---the speed and memory capacity to finally enable human-level AI---will come in the form of quantum computers \cite{Musser2018}. 

However, a growing cadre of researchers is questioning the basis of the ``all in the brain'' information-processing model for understanding intelligence and for creating AI.  Writing about what he calls ``The cul de sac of the computational metaphor,'' computer scientist Rod Brooks argues, ``The reason for why we got stuck in this cul-de-sac for so long was because Moore's law just kept feeding us, and we kept thinking, `Oh, we're making progress, we're making progress, we're making progress.' But maybe we haven't been'' \cite{Edge2019}.  In fact, a number of cognitive scientists have argued for decades for the centrality of the body in all cognitive activities.  One prominent proponent of these ideas, the psychologist Mark Johnson, writes of a research program on \textit{embodied cognition}, gaining steam in the mid-1970s, that ``began to provide converging evidence for the central role of our brains and bodies in everything we experience, think, and do'' \cite{Johnson2017}. Psychologist Rebecca Fincher-Kiefer characterizes the embodied cognition paradigm this way:  ``Embodied cognition means that the representation of conceptual knowledge is dependent on the body: it is multimodal..., not amodal, symbolic, or abstract.  This theory suggests that our thoughts are grounded, or inextricably associated with, perception, action, and emotion, and that our brain and body work together to have cognition'' \cite{Fincher2019}.

The evidence for embodied cognition comes from a diverse set of disciplines.  Research in neuroscience suggests, for example, that the neural structures controlling cognition are richly linked to those controlling sensory and motor systems, and that abstract thinking exploits body-based neural ``maps'' \cite{Epstein2017}.  As neuroscientist Don Tucker noted, ``There are no brain parts for disembodied cognition'' \cite{Tucker2007}. Results from cognitive psychology and linguistics indicate that many, if not all, of our abstract concepts are grounded in physical, body-based internal models \cite{Barsalou2005}, revealed in part by the systems of physically based metaphors found in everyday language \cite{Lakoff2008}.

Several other disciplines, such as developmental psychology, add to evidence for embodied cognition.  However, research in AI has mostly ignored these results, though there is a small group of researchers exploring these ideas in subareas known as ``embodied AI,'' ``developmental robotics,'' ``grounded language understanding,'' among others. 

Related to the theory of embodied cognition is the idea that the emotions and the ``irrational'' biases that go along with our deeply social lives---typically thought of as separate from intelligence, or as getting in the way of rationality---are actually key to what makes intelligence possible.  AI is often thought of as aiming at a kind of ``pure intelligence,'' one that is independent of emotions, irrationality, and constraints of the body such as the need to eat and sleep.  This assumption of the possibility of a purely rational intelligence can lead to lurid predictions about the risks we will face from future ``superintelligent'' machines.  

For example, the philosopher Nick Bostrom asserts that a system's intelligence and its goals are orthogonal; he argues that ``any level of intelligence could be combined with any final goal'' \cite{Bostrom2014}. As an example, Bostrom imagines a hypothetical superintelligent AI system whose sole objective is to produce paperclips; this imaginary system's superintelligence enables the invention of ingenious ways to produce paperclips, and uses up all of the Earth's resources in doing so.   

AI researcher Stuart Russell concurs with Bostrom on the orthogonality of intelligence and goals. ``It is easy to imagine that a general-purpose intelligent system could be given more or less any objective 
to pursue, including maximizing the number of paper clips or the number of known digits of pi''
\cite{Russell2019b}. Russell worries about the possible outcomes of employing such a superintelligence to solve humanity's problems:  ``What if a superintelligent climate control system, given the job of restoring carbon dioxide concentrations to preindustrial levels, believes the solution is to reduce the human population to zero?...If we insert the wrong objective into the machine and it is more intelligent than us, we lose'' \cite{Russell2019c}.

The thought experiments proposed by Bostrom and Russell seem to assume that an AI system could be ``superintelligent'' without any basic humanlike common sense, yet while seamlessly preserving the speed, precision and programmability of a computer.  But these speculations about superhuman AI are plagued by flawed intuitions about the nature of intelligence.  Nothing in our knowledge of psychology or neuroscience supports the possibility that ``pure rationality'' is separable from the emotions and cultural biases that shape our cognition and our objectives.  Instead, what we've learned from research in embodied cognition is that human intelligence seems to be a strongly integrated system with closely interconnected attributes, including emotions, desires, a strong sense of selfhood and autonomy, and a commonsense understanding of the world.  It's not at all clear that these attributes can be separated.

\section*{Conclusions}

The four fallacies I have described reveal flaws in our conceptualizations of the current state of AI and our limited intuitions about the nature of intelligence.  I have argued that these fallacies are at least in part why capturing humanlike intelligence in machines always turns out to be harder than we think.

These fallacies raise several questions for AI researchers. How can we assess actual progress toward ``general'' or ``human-level'' AI?  How can we assess the \textit{difficulty} of a particular domain for AI as compared with humans?  How should we describe the actual abilities of AI systems without fooling ourselves and others with wishful mnemonics?  To what extent can the various dimensions of human cognition (including cognitive biases, emotions, objectives, and embodiment) be disentangled?  How can we improve our intuitions about what intelligence is?

These questions remain open.  It's clear that to make and assess progress in AI more effectively, we will need to develop a better vocabulary for talking about what machines can do.  And more generally, we will need a better scientific understanding of intelligence as it manifests in different systems in nature.  This will require AI researchers to engage more deeply with other scientific disciplines that study intelligence.  

The notion of \textit{common sense} is one aspect of intelligence that has recently been driving collaborations between AI researchers and cognitive scientists from several other disciplines, particularly cognitive development (e.g., see \cite{DARPA}).  There have been many attempts in the history of AI to give humanlike common sense to machines\footnote{Some have questioned why we need machines to have   \textit{humanlike} cognition, but if we want machines to work with   us in our human world, we will need them to have the same basic   knowledge about the world that is the foundation of our own   thinking.}, ranging from the logic-based approaches of John McCarthy \cite{McCarthy1986} and Douglas Lenat \cite{Lenat1990} to today's deep-learning-based approaches (e.g., \cite{Zellers2019}).  ``Common sense'' is what AI researcher Oren Etzioni called ``the dark matter of artificial intelligence,'' noting ``It's a little bit ineffable, but you see its effects on everything'' \cite{Knight2018}. The term has become a kind of umbrella for what's missing from today's state-of-the-art AI systems \cite{Davis2015,Levesque2017}.  While common sense includes the vast amount of knowledge we humans have about the world, it also requires being able to use that knowledge to recognize and make predictions about the situations we encounter, and to guide our actions in those situations.  Giving machines common sense will require imbuing them with the very basic ``core,'' perhaps innate, knowledge that human infants possess about space, time, causality, and the nature of inanimate objects and other living agents \cite{Spelke2007}, the ability to abstract from particulars to general concepts, and to make analogies from prior experience.  No one yet knows how to capture such knowledge or abilities in machines. This is the current frontier of AI research, and one encouraging way forward is to tap into what's known about the development of these abilities in young children.  Interestingly, this was the approach recommended by Alan Turing in his 1950 paper that introduced the Turing test.  Turing asks, ``Instead of trying to produce a programme to simulate the adult mind, why not rather try to produce one which simulates the child's?'' \cite{Turing1950}

In 1892, the psychologist William James said of psychology at the time, ``This is no science; it is only the hope of a science'' \cite{James1892}.  This is a perfect characterization of today's AI. Indeed, several researchers have made analogies between AI and the medieval practice of alchemy.  In 1977, AI researcher Terry Winograd wrote, ``In some ways [AI] is akin to medieval alchemy.  We are at the stage of pouring together different combinations of substances and seeing what happens, not yet having developed satisfactory theories...but...it was the practical experience and curiosity of the alchemists which provided the wealth of data from which a scientific theory of chemistry could be developed'' \cite{Winograd1977}.  Four decades later, Eric Horvitz, director of Microsoft Research, concurred: ``Right now, what we are doing is not a science but a kind of alchemy'' \cite{Metz2017}.  In order to understand the nature of true progress in AI, and in particular, why it is harder than we think, we need to move from alchemy to developing a scientific understanding of intelligence.

\pagebreak

\section*{Acknowledgments}  
This material is based upon work supported by the National Science Foundation under Grant No.\ 2020103.  Any opinions, findings, and conclusions or recommendations expressed in this material are those of the author and do not necessarily reflect the views of the National Science Foundation. This work was also supported by the Santa Fe Institute.  I am grateful to Philip Ball, Rodney Brooks, Daniel Dennett, Stephanie Forrest, Douglas Hofstadter, Tyler Millhouse, and Jacob Springer for comments on an earlier draft of this manuscript.  

\bibliography{MD.bib}
\end{document}